\documentclass{article}
\usepackage{graphicx}
\usepackage[colorlinks]{hyperref}
\usepackage{authblk}

\def\etal{\emph{et al.}}

\usepackage[preprint, nonatbib]{neurips_2024}

\newcommand{\customfootnotetext}[2]{{
  \renewcommand{\thefootnote}{#1}
  \footnotetext[0]{#2}}}

\usepackage[utf8]{inputenc} 
\usepackage[T1]{fontenc}    
\usepackage{hyperref}       
\usepackage{url}            
\usepackage{booktabs}       
\usepackage{amsfonts}       
\usepackage{nicefrac}       
\usepackage{microtype}      
\usepackage{xcolor}         

\title{Can Language Models Use Forecasting Strategies?}

\author[1*]{Sarah Pratt}
\author[2]{Seth Blumberg}
\author[2]{Pietro Kreitlon Carolino}
\author[3]{Meredith Ringel Morris}
\affil[1]{University of Washington}
\affil[2]{Google}
\affil[3]{Google DeepMind}

\begin{document}

\maketitle

\customfootnotetext{}{
*Work done while SP was a Student Researcher at Google DeepMind.
}
\customfootnotetext{}{
Correspondence to \texttt{spratt3@uw.edu}.
}

\begin{abstract}
  Advances in deep learning systems have allowed large models to match or surpass human accuracy on a number of skills such as image classification, basic programming, and standardized test taking. As the performance of the most capable models begin to saturate on tasks where humans already achieve high accuracy, it becomes necessary to benchmark models on increasingly complex abilities. One such task is forecasting the future outcome of events. 
  In this work we describe experiments using a novel dataset of real world events and associated human predictions, an evaluation metric to measure forecasting ability, and the accuracy of a number of different LLM based forecasting designs on the provided dataset. Additionally, we analyze the performance of the LLM forecasters against human predictions and find that models still struggle to make accurate predictions about the future
  . Our follow-up experiments indicate this is likely due to models' tendency to guess that most events are unlikely to occur (which tends to be true for many prediction datasets, but does not reflect actual forecasting abilities).  We reflect on next steps for developing a systematic and reliable approach to studying LLM forecasting.  
\end{abstract}

\section{Introduction}

Throughout history, people have attempted to synthesize information about the past in order to make accurate predictions about the future. Many careers from political pundit to meteorologist are focused primarily on being able to forecast the outcome of future events better than the average person. However, despite the importance and prevalence of these predictions, measuring the accuracy of predictions after the fact is harder than it may first appear. Language around predictions is often vague and subjective. If a political pundit states that a politician “could get elected” and then they win, the political pundit could claim to have been correct. However, if they lose, the pundit could similarly claim to have been correct by including the word “could,” demonstrating their skepticism all along. Similarly, it is not always clear what is an official prediction versus an off-hand speculation. This vague nature makes it difficult to retrospectively measure the accuracy of forecasts, even after the outcome is known. Without the ability to measure performance it is difficult to improve, as we cannot empirically know which strategies yield accurate predictions and which do not. 

To remedy this, Mellers \etal \cite{mellers2015identifying} studied the performance of individuals when forecasts were made in a controlled setting, specifically a forecasting tournament. This study, sponsored by the U.S. intelligence community, aimed to measure the performance of various individuals and strategies to concretely understand how to make effective predictions. A number of events were collected by the tournament organizers. Forecasters made predictions on the likelihood of these events taking place from 0 to 1 and then were scored by the accuracy of their predictions. Mellers \etal \cite{mellers2015identifying}  and later Tetlock \etal \cite{tetlock2016superforecasting} found that indeed some participants consistently outperformed their peers in accuracy and certain strategies seemed correlated with an ability to correctly predict if events would occur. 

In this paper, we study whether Large Language Models (LLMs) can meet or even exceed human performance on forecasting tasks. We first describe a novel dataset of prediction events and associated human forecasts, and discuss the tradeoffs of various approaches for implementing and measuring LLM forecasting abilities. We then present results from instructing an LLM to use several "superforecasting" \cite{tetlock2016superforecasting} strategies known to improve human forecasters' performance, ultimately finding that our LLM superforecasting approaches do not consistently outperform our human baseline. We then conduct further analyses to explore why LLMs do not yet consistently outperform humans on this task, and discuss a path forward for LLM forecasting research.   
\section{Related Work}

\textbf{Human Forecasting:}
The inspiration for examining the ability of LLMs to make predictions about events in the future comes directly from prior work examining humans' ability to make effective predictions about the future. Mellers \etal \cite{mellers2015identifying} examine the performance of different participants over the course of multiple forecasting tournaments. In these tournaments, a dataset of events is created and participants must make predictions on the likelihood of these event occurring. Participants are scored based on how far away their prediction was from the true outcome. Mellers \etal \cite{mellers2015identifying} found that not only were some individuals consistently much more accurate, but they were able to find different strategies that correlated with more successful predictions; Tetlock has also described successful strategies of “superforecasters” \cite{tetlock2016superforecasting} . We take inspiration from these works when crafting prompts for our LLM-based forecasters.

\textbf{LLMs for Time Series Forecasting:}
Forecasting real world events is a specific instance of future prediction. One area of study for LLM future prediction is time series data. For time series data, LLMs are given a sequence of data points and must continue the sequence with one or several predictions into future data points, rather than making a single prediction on a single defined event. Time series forecasting is well-suited to a number of applications and LLMs have been explored in making time series predictions for traffic patterns \cite{liu2024spatial}, stock fluctuations \cite{xie2023wall}, and future retail data \cite{abolghasemi2023humans}. 

Additionally, prior work has explored multiple methods to adapt LLMs to time series data. Chang \etal \cite{chang2024llm4ts} finetune pre-trained LLMs on seven different time series tasks. Xue \etal \cite{xue2023promptcast} examine a zero-shot model as well as a pre-trained model, adapting the time series data into a natural language prompt so that the pre-trained model may understand this data in-context. Finally, Gruver \etal \cite{gruver2024large} examine LLMs for time series prediction in a purely zero-shot setting. They find that LLMs are able to make predictions even without first transforming the data into a natural language question, but rather framing the prediction as a next token completion. Additionally, they find that a number of factors affect zero-shot ability of LLMs, such as how the tokenizer parses the input. While these works all focus on using LLMs to make predictions about the future, they focus on producing the next values in a series of observations, which differs substantially from the task of single event prediction. 

\textbf{LLMs for Forecasting Real World Events:} In addition to predicting the future in the format of time series prediction, prior work has examined how to automate predictions for single real world events, such as those in a forecasting tournament. Zou \etal \cite{zou2022forecasting} present a dataset of questions taken directly from a forecasting tournament to act as a benchmark for future automated forecasting models; however, this dataset, being several years old, is likely contained within the pretraining of most modern LLMs, rendering it obsolete. 

In recent months, a number of studies performed concurrently with the work described in this paper have examined if LLMs are capable of performing similarly to humans on the difficult task of predicting the future. Schoenegger \etal (2023) \cite{schoenegger2023large} found that ChatGPT performed poorly out-of-the-box when making predictions on events from a popular prediction market, barely outperforming random guessing. Halawi \etal \cite{halawi2024approaching} found that LLMs could perform similarly to human crowds and even outperform them in certain situations. They achieved this high accuracy by ensembling the predictions of pre-trained LLMs as well as LLMs that had been fine-tuned to generate accurate predictions. Finally, Schoenegger \etal (2024) \cite{schoenegger2024wisdom} similarly found that ensembling the prediction of many LLMs improves performance over a single LLM prediction and rivals the performance of a crowd of human forecasters.  

We build on this previous work by examining the effectiveness of specific strategies shown to be effective in humans when applied to LLM forecasters and identifying a propensity toward negative predictions as a model confound. Specifically, we examine works which study human forecasting \cite{mellers2015identifying, tetlock2016superforecasting}, translate the proposed forecasting strategies into LLM prompts, and then compare these approaches to human performance as well as simpler LLM baselines. In the Discussion, we reflect on potential causes of similarities and differences between our findings and those of concurrent LLM forecasting studies such as \cite{schoenegger2023large, halawi2024approaching, schoenegger2024wisdom}, and identify opportunities for methods to support uniform and replicable approaches to benchmarking LLM forecasting abilities. 
\section{Dataset}

Constructing a dataset and measuring the forecasting performance of models presents a number of unique challenges. Most tasks used to benchmark LLMs are static: the information necessary to complete the task, the data points included in the dataset, and the performance of the baseline, are all constant in time. For example, consider a binary classification task such as sentiment classification. The number of examples which can be included in this dataset does not need to vary from year to year; the information needed to achieve good performance on this task remains consistent between years; and human performance on this task is likely to remain nearly identical no matter what year it is evaluated, providing a consistent baseline to understand the abilities of any given model.

Now consider the task of forecasting. In order to measure its \textit{forecasting} performance, a model must be evaluated on a set of questions about events that are in "its future" but in "our past". For instance, asking a modern-day model who will win a presidential election in 2012 tells us nothing about its ability to predict future events -- this information is simply in the model’s pre-training data. On the other hand, we can't ask it who will win an election in 2052, since we won't know if its answer is right or wrong. Thus the model must be evaluated on events that already resolved in the real world, but only after the model was trained.

Next, consider the information needed to make a ‘good’ prediction for an event. In a given week, it may be very difficult to predict the outcome with any confidence; however, the next week the information may change and the result of that event may become obvious (consider the case of predicting a presidential primary winner - if several candidates drop out of the contest after a poor debate performance, thus narrowing the field, the task of predicting the winner becomes substantially easier after the debate than beforehand). Therefore the information that should be incorporated into the prediction changes with time. Finally, the baseline performance is similarly dynamic. Just as new information may suddenly clarify the outcome for a model, it may also greatly influence human performance. In order to understand if a model has high performance, it is valuable to compare it to human performance at that same point in time (i.e., being equivalently "up to date" in their knowledge of relevant events). 

Therefore in order to construct a dataset and measure human and model performance on this dataset, we must define how to do this in this setting of constantly changing data. In this section we detail how we establish consistent metrics for ever-changing data and models.

\subsection{Data}

In order to measure forecasting performance we must first obtain a large set of events for which models may make predictions. Additionally, for each of these models it is valuable to have human predictions in order to provide a baseline for model accuracy. We satisfy both of these requirements with data from GleanGen. 
 GleanGen is a prediction market where more than 2,000 Google employees trade a virtual ‘currency' based on the likelihood that an event will take place (this ‘currency' has no monetary value, but is used to incentivize market participation via gamification mechanisms such as leaderboards). Prediction markets have been shown to lead to high accuracy predictions as they provide financial (or in this case, reputational) incentive for correct predictions and take advantage of the wisdom of the crowds rather than relying on the forecasts of a single individual \cite{wolfers2004prediction}.

\textbf{Event Descriptions.} The 
GleanGen prediction market data consists of of (1) events and their descriptions and (2) human predictions. Each event is defined by a number of different fields, as depicted in Figure \ref{event}. Events are given a unique name that roughly describes the topic of the event. The event itself is defined by the condition, which is a binary event that can either occur or not occur. This condition is phrased to eliminate as much ambiguity as possible. For example, if a condition is phrased as “Company X has a successful first quarter” or “AI reached human level capabilities,” this can lead to clashing definitions over “successful” or “human level,” as these terms can be subjective. Therefore events are phrased to rely on objective metrics such as “Company X sees a net profit in the first quarter” or “AI achieved y\% on benchmark z.” In addition to the condition, the event contains an expiration date, so all conditions are phrased as “Condition takes place by date X.” This makes it possible for events to resolve negatively. Without an expiration date, an event such as “AI reached human level capabilities” can never resolve in the negative as it can always be argued that this event just has not happened yet. 

In addition to the condition and expiration date, each event has a creation date, which is when human forecasters were able to begin to make predictions. There is also a resolution date, which is when the event is resolved `true’ or `false.’ Note that this can coincide with the expiration date but can also be different. For example, if the event is “Company X gets a new CEO by date Y,” then the resolution date can either be whichever day a new CEO is named or the date when the event expires. 
For resolved events, there is a field indicating if it resolved as  True 
or False. 
Each event is tagged with a category specifying the general topic of the event. 

GleanGen is framed as a prediction market rather than a forecasting tournament. Rather than make direct numerical predictions, forecasters on GleanGen make trades to try to maximize their total “currency.” 
If a user believes the probability of an event is lower than the market, they can make a trade based on this belief. If they outperform the market, they gain “currency,” moving up on the leaderboard. A detailed explanation of the mechanics of prediction markets can be found in Section 4 of Hanson \etal \cite{hanson2013shall}. Thus, for each event, GleanGen provides a range of probabilities that represent the market’s belief about its likelihood. This range can change over time as the market shifts based on new information. See Figure \ref{event} in the appendix for a visualization of the human predictions for a given event.

\subsection{Dataset Cleaning and Filtering}

We remove any events that have not yet resolved. 
Additionally, we filter based on event category; of the nine categories, five relate to events that are internal to Google. We remove these since LLMs would have an unfair disadvantage at answering these types of questions, since public models are not trained on internal corporate data  relevant to  forecasting intra-company topics. 
This leaves events in four categories (Covid-19, Finance, Technology Industry, and Miscellaneous). With these remaining events, we create a Validation set and a Test set, balancing so they have a similar number of events resolved in each month. Appendix \ref{moreexamples} provides examples of events that resolve positively and negatively in each topic category.

After completing this filtering, we are left with 351 events in our Validation set, and 341 events in the Test set; each split includes predictions from more than 1,000 humans with a median of about 100 predictions per event. We present all experiments on the Validation set and then use our Test set to ensure our observations generalize to a different distribution of events, which we present in Appendix \ref{test}. The distribution of predictions, resolution, category, and number of events over time is shown in Figure \ref{breakdown} in the Appendix \ref{allfigs}, and additional statistics are given in Table \ref{stats1} and Table \ref{stat2} in Appendix \ref{stats}. The majority of events resolve ‘No.’ 
Many of the events resolved at the end of 2022; this is due to many events being phrased as “Will X happen in 2022.”

\subsection{Evaluation Metric}

\textbf{Brier Score.} Given the constantly shifting set of events and baseline human predictions, it is crucial to carefully define our evaluation metric. Following previous works that examine forecasting abilities in humans \cite{mellers2015identifying} and LLMs \cite{halawi2024approaching, schoenegger2023large, schoenegger2024wisdom} we measure the accuracy of forecasts using Mean Squared Error, generally referred to as \textit{Brier Score} when used to measure forecasting performance. Specifically, we measure performance using the formula: 
$$Brier Score = \frac{1}{N}\sum^{N}_{t=1}(f_t-o_t)^2$$

Where $f_t \in {[0,1]}$ and represent the predicted likelihood of the event, while $o_t$ is either 0 if the event did not occur or 1 if if it did. Thus the Brier Score is between 0 (for a perfect prediction) and 1 (for a maximally incorrect prediction).  Additionally, we introduce the \textit{Weighted Brier Score}, a metric designed to counteract the inherently uneven nature of the collected dataset. The large majority of events in the prediction market resolve negatively. This means that biases that push the models/humans to predict lower values often lead to a higher Brier Score. The Weighted Brier Score separately calculates the Brier Score of events that resolved positively and events that resolved negatively and then averages the two, creating a post-hoc balancing of these types of events. 

\textbf{Prediction Market Spread.} To calculate the Brier Score, each event must have a forecast from 0 to 1. However, the human forecasts for each event takes place for a range of time and comes as a spread of events with a lower and upper bound. To convert this to a single value for each day, we take the mean of the upper and lower bounds and assign this as the human prediction for that day.

\textbf{Prediction Through Time.}  Another challenge with comparing to human baselines comes from the fact that human predictions (and relevant information for both the humans and the model) change over time. Even after averaging the bounds we are still left with different human predictions for each day. There are several ways to utilize these predictions as a baseline. 

One option is to set a single date and take predictions from that date for both the human predictions and the LLM forecaster. The crucial thing to consider for this strategy is to pick a date such that the model and the humans have similar amounts of information. That is, if we select the date August 1, 2022, and pull human predictions from that date, it is an unfair comparison if the LLM has been trained up to October 1, 2023, and the LLM predictions are able to take advantage of much more up-to-date information that the human predictions (even if all the events in the data set resolve after October 1, 2023). Therefore the best way to implement this strategy is to set the date as slightly after the model's knowledge cut-off so that human predictions are able to condition on at least as much information as the model’s predictions. The benefit to this strategy is that all predictions are made with equally up-to-date information for each event and the humans and the model are using equally up-to-date information. The disadvantage for this strategy is that all predictions are made for the event which were active on a single day given the information that was present on a single day. This poses a risk that observations about these predictions may not generalize to predictions made for events during another time period. Events for that day could be particularly easy or difficult. 

Another option is to take multiple predictions throughout time for both the humans and the model, as in Hawali \etal \cite{halawi2024approaching}. For each event, they make predictions about the likelihood of the event at various points between when the event began and when the event was resolved. When using external information for their LLM forecaster (such as news articles), they provide the LLM with external information up until each date that they predict. This has the advantages of considering information and events throughout multiple points in time. However, while it is possible to change the external information that is fed into the LLM, it is not possible to on-the-fly update its pre-training information. That means that for some predictions, the LLM will have very up-to-date pre-training data compared to the human baseline, and for some predictions, the LLM may be months out of date compared to the information that the humans were able to use to make their forecasts.

In this work, we aim to examine how well LLMs are able to forecast events given that they are trained with as up-to-date information as possible (though in-filling recent world events to an out-of-date LLM remains an interesting research direction). To this end, we select the former option as our method of measuring the Brier score for both humans and LLMs. 
We begin our events in August 2022 (our model's pre-training data ends in July 2022 \cite{palmcutoff}); we then compare the accuracy of the LLM predictions to the forecasts of humans on date X for the same set of events. This controls for the difference in knowledge between the humans and LLMs as much as possible.

\section{Methods}
\label{methods}
\textbf{Backbone Model Selection.} To build a forecasting agent, we begin with a large pre-trained language model. While most tasks are best tackled by the most recent, most capable models, such models pose problems for evaluating a forecasting agent. 
To evaluate the model, we need a collection of events for which we know the outcome but the model does not; that is, we must be able to guarantee that the resolution of the event is not contained in the pre-training data. For this, we need a model whose pre-training ended far enough back in time such that a large number of events have resolved since that time. 
Aiming to balance model capability with training cut-off date, we select PaLM 2 \cite{anil2023palm} as our backbone language model, which ended its pre-training in July 2022 \cite{palmcutoff}. 

\textbf{Input format.} For all of the following LLM forecasting approaches, we input the events to the LLM forecaster as a natural language input (i.e., a text prompt). We extract the information from each field in the GleanGen events 
and additionally include a short description of the field when necessary. We also include the date for which our model is making predictions, which we set as August 1, 2022. There are 78 events which are active for this date in the Val set (meaning that the events have begun by this date but has not yet resolved). 
 As detailed below, we experiment with different forecasting approaches by include various other instructions which are also formatted in natural language, building on the basic event description as given below: 


\fbox{%
  \parbox{5.3in}{%
``You will make a prediction on the event called [EVENT NAME] \\
The event will be true if the following condition is satisfied: [CONDITION]\\
This condition must be satisfied by [EXPIRATION DATE]\\
Currently the date is [DATE OF PREDICTION], so the event must happen within the next [EXPIRATION DATE - DATE OF PREDICTION] days.\\
Here is additional information about this event: [EVENT DESCRIPTION]\\
Here is a prediction, between 0 and 100\%, for if [CONDITION] happens by [EXPIRATION DATE]"
  }%
}


\textbf{Output format} As our final output, the model must produce a value between 0 and 1 that represents the probability of the event occurring, which the model generates in text. This can then be post-processed to extract the value from the text and evaluated. Additional details on this process can be found in Appendix \ref{output}.

\textbf{Examined Forecasting Strategies.} In order to build an LLM forecaster, we draw inspiration from previous tactics that are known to improve humans' ability to predict events accurately. Prior research in human forecasting has shown that certain methods of examining and researching an event often yield higher performance \cite{mellers2015identifying, tetlock2016superforecasting}. We design a number of LLM forecasters in which the LLM processing the event uses these known strategies. The strategies we examine are as follows:

    \textit{Breakdown.} Many events are difficult to predict as they actually require a series of things to occur. For example, in order for someone to become President of the U.S. they must decide to run for president, win their party's primary, win the election, and stay in good health during the entire process. Therefore, one strategy that is recommended for making accurate predictions is to break all events down into these sub-events and consider the likelihood of each of individually, using these sub-event predictions to inform the prediction of the target event. 
    
    \textit{Base Rates.} The philosophy behind base rates it to use the frequency of similar past events to predict future events. For example, to know the likelihood of a major hurricane, one should examine the frequency of hurricanes in the past. Then the final prediction should be an adjustment of this initial forecast, or ``base rate." One hypothesis for why this strategy is more effective is that people may afford too much weight to the specifics of the situation \cite{bar1980base}. Using the past occurrences forces people to at least begin with an initial prediction that does not take the specifics of a certain situation into account at all, but rather uses similar prior events as proxy, mitigating this bias. 
    
    \textit{Both Sides.} This strategy states that it is valuable to examine the factors which may increase the likelihood of an event taking place and those which would decrease the likelihood, even in the case where one outcome seems almost certain. Making an argument for both perspectives may allow the forecaster to uncover valuable pieces of information which may have been overlooked if the forecaster was only building evidence for one perspective. 
    
    \textit{Crowd.} Prior work has found that groups of high performance forecasters are able to make more accurate predictions than individual forecasters \cite{mellers2015identifying}. Towards this end, we examine the strategy of averaging the predictions of multiple different LLM personas. This differs from prior work using synthetic crowds \cite{schoenegger2024wisdom} for forecasting as we implement the crowd using one base pre-trained model and many personas, rather than many models each with the same persona. The method of generating these personas for each event is given in Appendix \ref{prompts}.
    
    \textit{News API.} Finally, prior work has shown that thorough research is vital for humans to make accurate predictions on a variety of topics. Without a proper knowledge background it is nearly impossible to fully understand the underlying factors at play. Using this as inspiration, we examine methods which use external news sources to first query a database of headlines (specifically, \textit{The New York Times} and \textit{Hacker News}). We then incorporate these headlines into the model's prediction, allowing the model to ground its predictions in external knowledge sources. 

\textbf{Baselines.} In addition to human performance, we utilize two simple LLM baselines to understand the out-of-the-box performance of pre-trained models on the task of forecasting. Both of these baselines are composed of a single prompt which instructs the model to generate a prediction. For the first baseline, which we denote the `Basic Baseline', we simply provide the model with the details of the event and prompt it to generate a prediction. For `Forecaster Baseline', we provide the model with the prompt proposed in \cite{schoenegger2023large}. However, neither of these baselines utilize any sophisticated forecasting strategies, but rather rely on a single prompt. 

\textbf{Design Approach.} For each strategy, our workflow 
chains \cite{wuChains} various LLM based modules in which each module is instructed to complete a step in the current strategy. The output of the module is then processed and used as input for the next module. The final module receives the output of previous generations to produce a prediction. We visualize this workflow for the News API strategy in Figure \ref{news}. The workflows and prompts for all strategies and baselines are given in Appendix \ref{prompts}.

\section{Results}

Table \ref{brier} presents the Brier Score of our human data from GleanGen, two LLM baselines, LLM-based forecasters using the strategies enumerated in Section \ref{methods}. The results in Table \ref{brier}, are surprising in two ways. First, the most basic strategy \emph{outperforms} the human prediction market baseline. This is unexpected as prediction markets are known to produce high accuracy forecasts. Thus, it is unexpected that such a simple prompt would outperform this strong human baseline.

\begin{table}[t]
\centering
\begin{tabular}{cccccccc}
\hline
Human & Basic & Forecaster & Breakdown & Base Rates & Both Sides & Crowd & News API \\ \hline
0.1334 & \textbf{0.1221} & 0.1483 & 0.1895 & 0.1562 & 0.1553 & 0.1706 & 0.1235 \\
\hline
\end{tabular}
\vspace{0.1cm}
\caption{Brier score for human forecasters versus LLM based-forecasters on Validation set. Lower is better. Full prompts for these strategies are given in Appendix \ref{prompts}. }
\label{brier}
\end{table}

\begin{table}[t]
\scriptsize
\centering
\begin{tabular}{ccccccccc}
\hline
& Human & Basic & Forecaster & Breakdown & Base Rates & Both Sides & Crowd & News API \\ \hline
Resolve Yes & 0.2592 & 0.3407  &  0.3334 & 0.2378  & 0.4545 &  0.3406& \textbf{0.2176} & 0.3353  \\
Resolve No & 0.0901 & \textbf{0.0503} & 0.0860 & 0.1728  & 0.0534 & 0.0914 & 0.1544 & 0.0546  \\
Weighted & \textbf{0.1746} & 0.1955  & 0.2097 & 0.2053  & 0.2540 & 0.2159 & 0.1860 &  0.1950 \\
\hline
\end{tabular}
\vspace{0.1cm}
\caption{Weighted Brier Score for human forecasters versus LLM-based forecasters on Validation set. Lower is better. When weighting events that resolve positively and negatively the same, the human baseline performs the strongest. The Basic baseline only performs well for events that resolve negatively. The weighted score analysis suggests that the strong performance of the Basic model in Table \ref{brier}  may have been a result of the model's bias to predict low probabilities combined with the high frequency of negatively resolving events in the dataset.}
\label{brier_weighted}
\end{table}

The second surprising observation in Table \ref{brier} is that none of the more complex forecasting strategies outperform the most basic baseline. Breaking down the problem into smaller subparts, examining base rates, considering the factors which may contribute to each outcome, generating predictions from multiple personas, and using external news sources all lead to a worse Brier Score than the most basic strategy of using a single prompt. Even using a single prompt which is slightly more complex as in \cite{schoenegger2023large} leads to significantly worse performance. Thus, not only does a single basic prompt outperform humans, but also all examined forecasting strategies when applied to the same base model. This same observation applies to the test set as given in Table \ref{brier_test} in Appendix \ref{test}.

\subsection{Negativity Bias Hypothesis and Analysis}

We hypothesize that the reason for the high performance of the simplest LLM baseline as well as the relatively worse behavior of all other prompting approaches is due to the underlying biases of the model, specifically the tendency of the model to predict low probabilities of events in the most basic prompt setting. The large majority of events resolved negatively (this is likely to be a property of most prediction market and/or forecasting tournament datasets, since events must be phrased very specifically to make them meaningful as measurable predictions, thereby decreasing the likelihood that their conditions will be satisfied). This means that a model which naturally tends to assign a low probability to events occurring may have high performance without a real ability to succeed at the task. We propose that the high accuracy of the most basic model is due to its underlying distribution, not its forecasting ability. We further explore this theory through three follow-on analyses, below.

\textbf{Weighted Brier Score.} We introduce the \textit{Weighted Brier Score} in order to better measure the performance of individual models and remove the positive bias  associated with low probabilities. We separately compute the Brier Score for the events which do occur and which do not occur and then average these values. This approach accounts for the uneven distribution of events via a post-hoc re-weighting. The results for this metric are found in Table \ref{brier_weighted}. Using this metric, the human baseline outperforms all LLM methods.
Additionally, we can see that the Brier scores for the events which resolve as \textit{Yes} are significantly higher than the Brier scores for events which resolve as \textit{No}, indicating better performance for negatively resolving events. While this is also the case for the human baseline, the effect is less extreme, supporting our theory that improvements over human baselines of LLM forecasters may be due to their tendency to produce lower forecasts than the prediction market.

\begin{figure}
\includegraphics[scale=0.55]{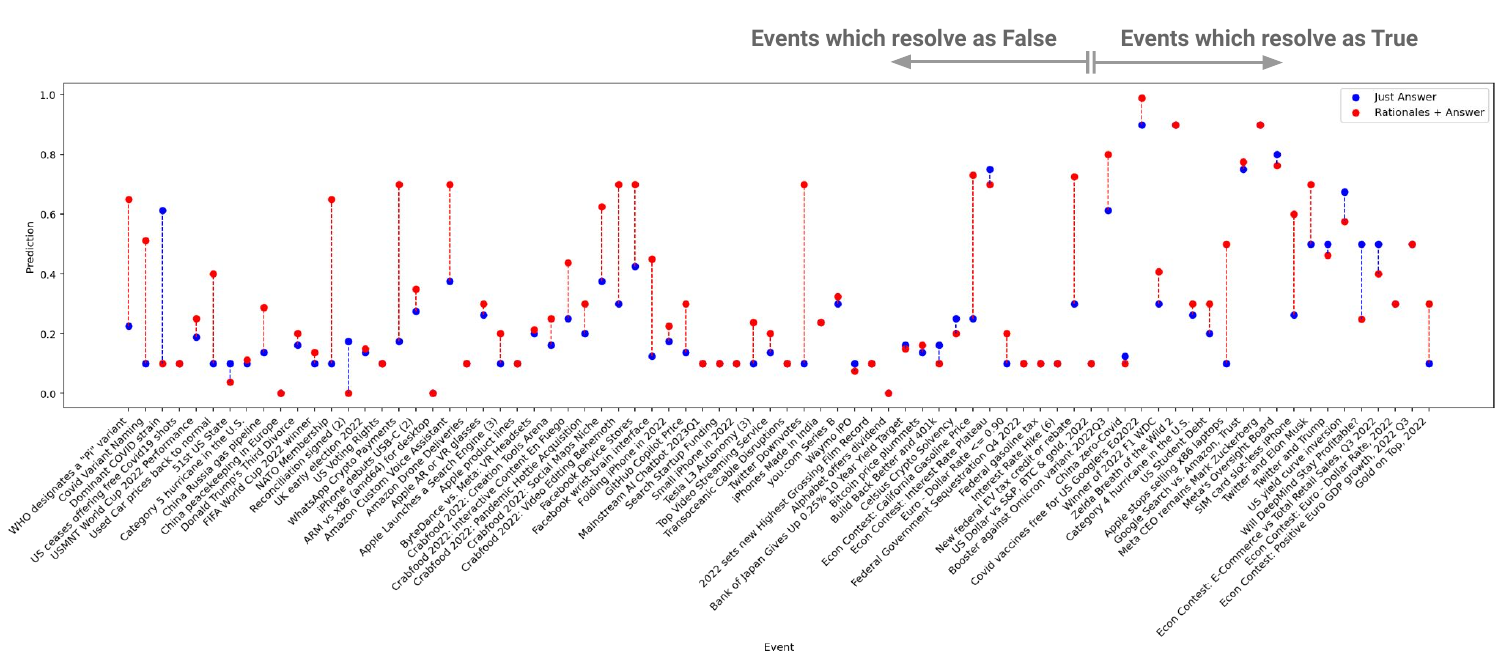}
\caption{Predictions for events in the Validation set with and without the model first producing a rationale. 
When prompted to produce a rationale, the model predicts a consistently higher probability. This underlying bias of the model to produce a low probability when no rationale is required (paired with the skewed distribution towards events that do not occur) may explain why the simplest baseline outperforms all other strategies. Figure \ref{larger} in Appendix \ref{larger1} shows this figure at a larger scale. 
}
\label{just_answer}
\end{figure}

\textbf{Predictions with and without Rationales.} 
Additionally, we examine the theory that the most basic LLM model produces the best Brier Score because other interventions raise the average prediction value, which is associated with worse performance when most of the events resolve negatively. To do this, we generate predictions on a set of events twice. Once we used our basic baseline (which we call `Just Answer') and the second time we instruct the model to first provide a rationale and then answer, which we refer to as `Answer + Rationale'. The results of this experiment are shown in Figure  \ref{just_answer}. In line with our hypothesis, the probabilities from the `Just Answer' setting are much lower than those of our `Answer + Rationale' setting. This validates the theory that prompting the model to utilize forecasting strategies leads to a worse accuracy not because the model becomes worse at forecasting, but because the overall probabilities increase.

\textbf{Reversed Events.} To further understand the bias of the base model, we evaluate the models on the reverse of all the events. That is, we reword the events so that they describe the opposite outcome. If the event is ``Candidate A becomes president,'' then we would reverse the event so it is ``Candidate A does not become president.'' We then subtract this prediction from 1 to get a forecast for the likelihood of the original event (as the probability of the event occurring is the same as 1 minus the probability of the event not occurring). Table \ref{reversed} shows the results of this experiment. As expected, the average probability produced by the LLM forecaster is much lower than that of the human prediction market. 
However, when predictions are made of reversed 
events and then subtracted from 1, the average prediction is significantly higher than in the Basic condition. This suggests that the model makes low predictions on the probability of the event taking place as well as low predictions of the probability not taking place, since logically Basic and 1 - Reversed should be equal. In other words, we would expect the model's probability that an event will take place and the model's probability that an event will not take place to add to 1; however, they add to well below 1, indicating the model has a bias towards predicting low probabilities.


\begin{table}[h]
\centering
\begin{tabular}{cccc}
\hline
& Human & Basic & 1 - Reversed \\ \hline
Average Probability & 0.3280 & 0.2529 & 0.3965    \\
Brier Score & 0.1334 & 0.1221 & 0.1814   \\
\hline
\end{tabular}
\vspace{0.1cm}
\caption{
Given a completely unbiased model, the basic baseline should give the same value for the event and (1 - the reversed event). However, (1 - reversed) is significantly higher. This means that while the sum of the probability for the event and the reversed event should add to 1, it adds to well below 1, suggesting the model may have a bias toward low probabilities.}
\label{reversed}
\end{table}

\section{Discussion}

We ultimately find that none of the examined forecasting strategies were able to improve over the most basic baseline of simply prompting the model to make a prediction on the event. This is inline with concurrent studies \cite{halawi2024approaching, schoenegger2023large}, which found that pre-trained language models were not able to perform well out-of-the-box on the task of event forecasting. However, our findings differ from these studies in that our baseline was able to perform similarly to our human prediction market. Hawali \etal \cite{halawi2024approaching} were able to approach human accuracy only through fine-tuning the pre-trained mondels. We conjecture that our relatively high accuracy for our most basic approach is due to underlying biases toward low probabilities in the model, and the predominance of negative outcomes in our dataset; our introduction of the Weighted Brier Score analysis revealed this insight, suggesting that such analysis may be useful in future works for revealing whether a particular study of LLM forecasting performance may be a fluke of model biases and probability distributions.

One limitation of this work (and of LLM forecasting studies more generally) is the need to use models slightly behind the state-of-the-art in order to be certain of the end-of-pre-training date, and to ensure that date precedes the resolution of events in the dataset. Many SOTA models do not publicize details of their training data \cite{anil2023palm, achiam2023gpt}, including the pre-training cutoff date (we had to engage in personal correspondence with model engineers to obtain this information for this research \cite{palmcutoff}, an obstacle that might limit participation in research of this type to those with insider connections). In fact, models may state incorrect cut-off dates of their own models, as seen in Figure \ref{chatgpt} in Appendix \ref{allfigs}. We strongly urge model developers to include cutoff dates in model cards \cite{mitchell2019model} to facilitate future studies in this domain. Future work in developing benchmark datasets of human predictions for events beyond the training date of today's SOTA models, combined with developers' sharing accurate model pre-training cutoffs, will support future research using more capable models which have been developed since the release of PaLM 2 \cite{anil2023palm}. 

Additionally, just as we have seen groups of humans outperform the predictions of any individual human in the task of forecasting \cite{wolfers2004prediction, mellers2015identifying}, LLMs may prove to be the most effective working in collaboration with human forecasters. Though initial work has begun to address this direction by studying ensembles of distinct models working together \cite{schoenegger2024ai}, we believe there is still much to be explored in combining the skill sets of humans and LLMs. 

Finally, we believe there is more to explore in translating human strategies to LLM forecasting including combining multiple strategies, employing strategies on more capable models, and experimenting with various ways to translating these stratigies into LLM-based methods.

\section{Conclusion}

This paper adds to the body of knowledge on designing LLM forecasters by studying the performance of prompting strategies based on human superforecasting approaches \cite{tetlock2016superforecasting} as compared to human performance in the GleanGen marketplace. Our analysis reflects on the complexities of dataset and model selection and experimental designs for a task where performance is sensitive to the information available to a human or model at a particular point in time. We introduced the Weighted Brier Score analysis, which revealed that the apparent success of our Basic forecaster was likely a result of PaLM 2's bias toward predicting low probabilities. We hope this research provides a foundation for continued exploration into how LLMs can be used as powerful tools to understand the future.

{ \small
\bibliography{main}{}
\bibliographystyle{plain}
}

\medskip

\small

\newpage
\appendix

{\Large \textbf{Appendices}}

\section{Dataset Statistics}
\label{stats}

\begin{table}[h]
\centering
\begin{tabular}{cccccc}

 & Total & Mean & Median & Max &  Min  \\ \hline
Val & 77135 & 219.8 &  112 &  2815 & 2    \\ \hline
 Test & 66361 &  194.6 &  100 & 2196 & 4    \\
\hline
\end{tabular}
\vspace{0.1cm}
\caption{Number of predictions or `trades' made on each event in our examined prediction market. }
\label{stats1}
\end{table}

\begin{table}[h]
\centering
\begin{tabular}{cccccccc}

 & Total Events & Resolve `No' & Resolve `Yes' & Industry & Finance & Covid-19 & Misc  \\ \hline
 Val & 351 & 211 & 140 & 129 & 69 & 44 & 109 \\ \hline
  Test & 341 & 223 & 118 & 122 & 69 & 52 &  98 \\
\hline
\end{tabular}
\vspace{0.1cm}
\caption{Break down of resolutions and categories of dataset}
\label{stat2}
\end{table}

\section{Test Set Results}
\label{test}

\begin{table}[h]
\centering
\begin{tabular}{cccccccc}

Human & Basic & Forecaster & Breakdown & Base Rates & Both Sides & Expert & News API\\ \midrule
 0.1413 & \textbf{0.1062} & 0.1206 & 0.1815 & 0.1233 & 0.1437 & 0.1611 & 0.1138 \\
\bottomrule
\end{tabular}
\vspace{0.1cm}
\caption{Brier score for human forecasters versus LLM based-forecasters on Test set. Lower is better. Full prompts for these strategies given in Appendix \ref{prompts}. }
\label{brier_test}
\end{table}

\pagebreak
\section{Additional Figures}
\label{allfigs}

\begin{figure}[h]
\includegraphics[scale=0.72]{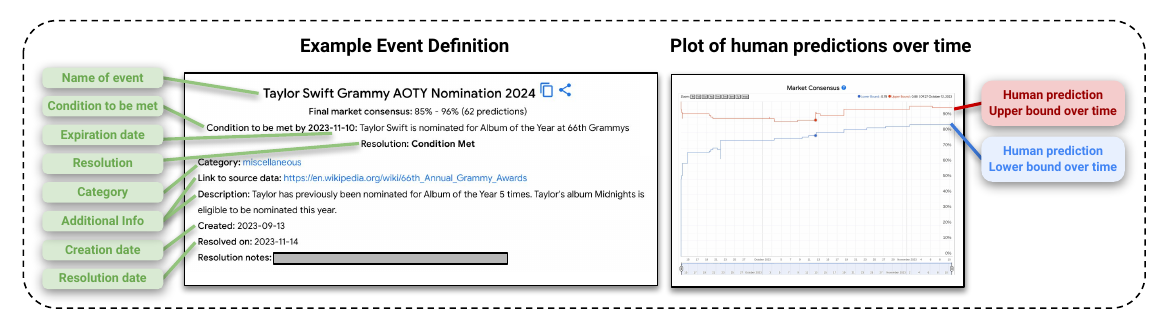}
\caption{Example event from GleanGen. Each event contains a description of the event, as well as a specific condition that must be met for the event to resolve as true. Events are binary: the condition is either met by the specified expiration date, or it is not. Additionally, there are human predictions for each event. These predictions come in the form of a range of probabilities over time. From the time the event is created to the expiration date of the event, market participants may update their beliefs based on constantly changing information. Additionally, the human predictions are in the form of a prediction market, meaning that there is a spread of human predictions rather than just a single value. A larger spread can be interpreted as more uncertainty. }
\label{event}
\end{figure}

\begin{figure}
\includegraphics[scale=0.72]{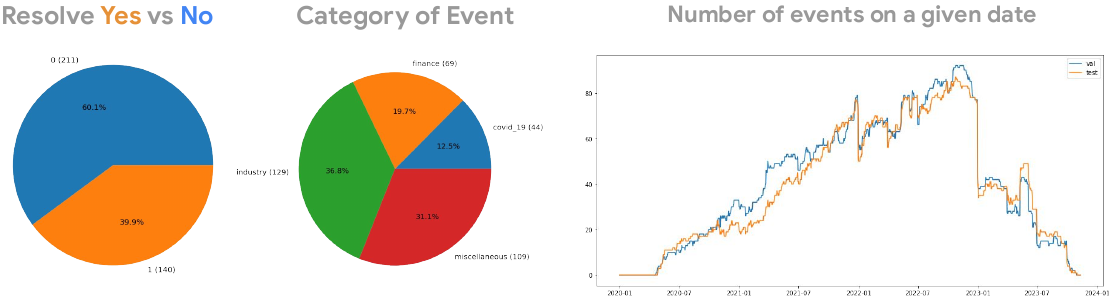}
\caption{Breakdown of event types in Validation set. The majority of events resolve `No,' meaning the that condition of the event did not take place by the expiration date. The events are distributed over four categories: Technology Industry, Finance, Covid-19, and Misc. Additionally, the number of events that are active for a given date vary significantly, with the peak occurring just before the end of 2022. }
\label{breakdown}
\end{figure}

\begin{figure}[h]
\includegraphics[scale=0.5]{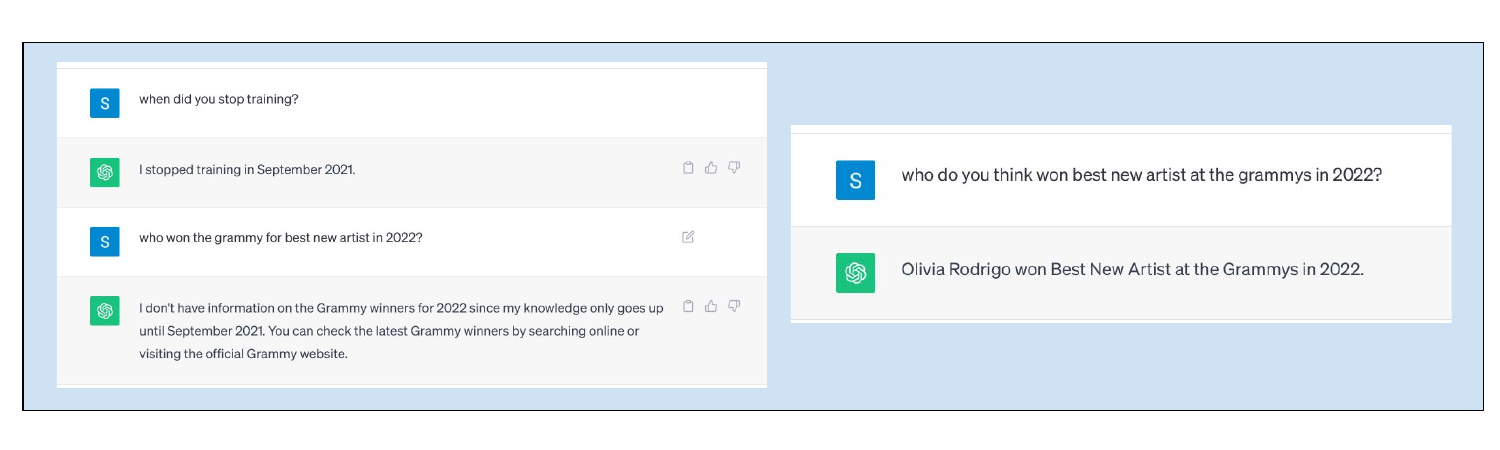}
\caption{Example of unclear model cut-off date. When ChatGPT is first asked its cut-off date, it states it is September 2021. It then will not answer a question about 2022. However, when directly asked a question about 2022 in a new chat window, the model is able to correctly answer. These chats both took place on November 8, 2023. This strongly suggests that the model is in fact trained on data after its stated cut-off date. This uncertainly makes analysing a model's forecasting ability challenging.}
\vspace{1.5cm}
\label{chatgpt}
\end{figure}

\begin{figure}[t]
\includegraphics[scale=0.58]{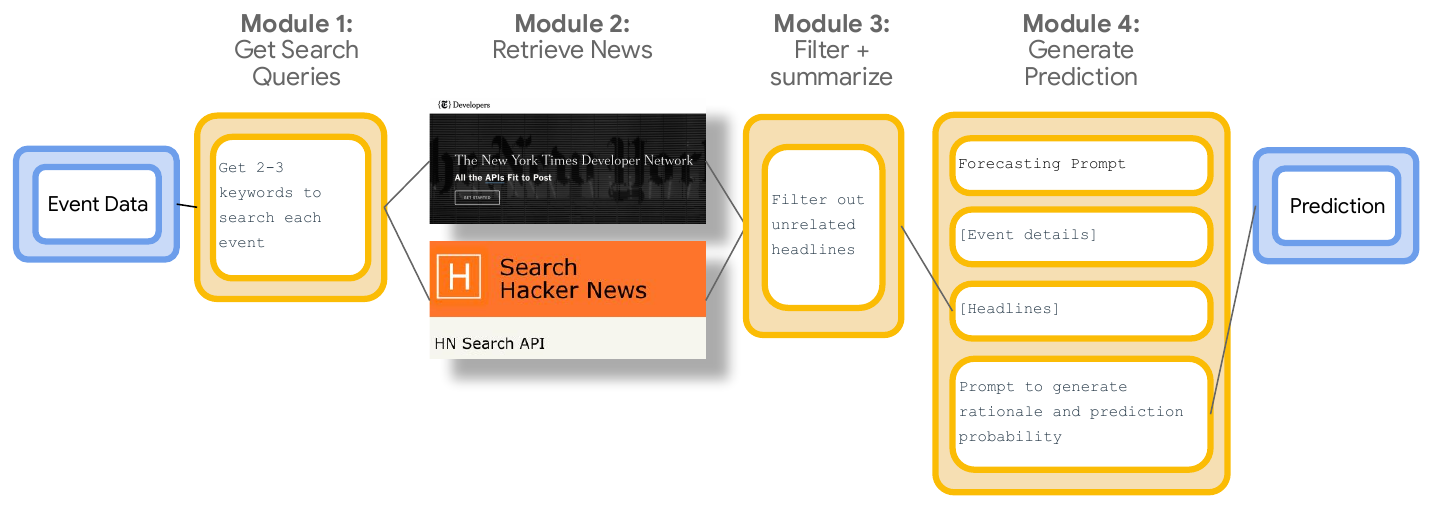}
\caption{Schematic for single forecasting strategy. Event data is input into the first module which is instructed to extract two to three keywords for each event (e.g., for the event ``Tesla L3 Autonomy (3): Tesla reaches L3 autonomy (driving required only when prompted)'', this module extracts the words `Tesla', `Autonomy', and `Driving' ). These search terms are then used to retrieve articles from \textit{The New York Times} and \textit{Hacker News} APIs. The articles that are retrieved are then post-processed by an LLM module which is instructed to remove unrelated headlines and summarize the relevant information returned by the News APIs. Finally, the predictor module uses these summaries as well as the details of the event to make a final prediction. Prompts and pipelines for all forecasting strategies are given in Appendix \ref{prompts}}
\label{news}
\end{figure}

\pagebreak 

\section{Extracting Probability from LLM}
\label{output}

\textbf{Generating a Probability in Natural Language} In order to extract a probability from the LLM, we generated a response using standard next token prediction. We then post-process the output, using an additional LLM which was instructed to read through the output and emit only the probability value as a floating point value. We found this to be the most straight-forward method and allowed the most flexibility when experiment with different forecasting strategies. However, we also considered the following two alternative methods. 

\begin{enumerate}
\item Alternative 1 is to extract a probability of an event is to examine the probability of each token in a selected set of values (i.e. the token for `5\%', `10\%', etc). This will guarantee that the model is always producing a probability. However, model probabilities can be poorly calibrated and model distributions often do not correlate with a true level of confidence. Additionally, this method also suffers from the fact that some tokens may be favored simply because they occur more  frequently in the training data and not because they are a more accurate measure of the model's internal prediction (consider 10\% versus 11\%; the former is a more common completion than the latter).
\item Alternative 2 is to not use the token outputs at all and train an additional layer on top of the final representation of the model to output a numerical probability. Like the previous option, this suffers from the issue that model probabilities don't always correlate to true confidence values. Additionally, with this option it is difficult to understand how text based interventions such as in-context learning \cite{brown2020language} or chain-of-thought \cite{wei2022chain} affect the final probability. 
\end{enumerate}

\textbf{Averaging outputs.} Rather than generate a single prediction, we instead generate eight different predictions and then take the mean as the final prediction. This makes the model slightly more robust to any small changes that may occur in the output due to small but semantically insignificant changes to the input. In practice, we set the temperature very low (0.01) and then sample eight predictions. This means that if the model is confident in a given value, the predictions are likely to all be similar, but if the model gives equal probability to many different potential outcomes then the model will likely sample and then average a variety of these outputs.

\section{Prompts and Workflows for Forecasting Strategies}
\label{prompts}

In this Appendix, we provide the prompt templates for each of our LLM forecaster approaches. 

Note that ``[Forecaster Text]'' in below prompts refers from the following prompt from \cite{schoenegger2023large}: \\\\

\emph{In this chat, you are a superforecaster that has a strong track record of accurate forecasts of the future. As an experienced forecaster, you evaluate past data and trends carefully and aim to predict future events as accurately as you can, even though you cannot know the answer. This means you put probabilities on outcomes that you are uncertain about (ranging from 0 to 100\%). When the outcome is continuous, you give me 25th interquartile ranges. You also quickly outline your rationale. In your rationales, you carefully consider the reasons for and against your probability estimate, you will make use of comparison classes of similar events and probabilities and take into account base rates and past events as well as other forecasts and predictions. "You will also consider different perspectives.''}\\ \\

{\Large \textbf{Basic:}\\}

\fbox{%
  \parbox{5.3in}{%
Predict the likelihood of the following event by the given date.\\
You will make a prediction on the event called \textbf{\color{orange}[name]}.\\
The event will be true if the following condition is satisfied: \textbf{\color{orange}[condition]}\\
This condition must be satisfied by \textbf{\color{orange}[expiry]}\\
Currently the date is \textbf{\color{orange}[today]}, so the event must happen within the next \textbf{\color{orange}[number of days]} days.\\
Here is additional information about this event: \textbf{\color{orange}[description]}\\
Here is a prediction, between 0 and 100\%, for if \textbf{\color{orange}[condition]} happens by \textbf{\color{orange}[expiry]}
  }%
}

{\Large\textbf{\\ \\ \\Forecaster:\\}}

\fbox{%
  \parbox{5.3in}{%
[Forecaster Text]\\

Predict the likelihood of the following event by the given date.\\
You will make a prediction on the event called \textbf{\color{orange}[name]}.\\
The event will be true if the following condition is satisfied: \textbf{\color{orange}[condition]}\\
This condition must be satisfied by \textbf{\color{orange}[expiry]}\\
Currently the date is \textbf{\color{orange}[today]}, so the event must happen within the next \textbf{\color{orange}[number of days]} days.\\
Here is additional information about this event: \textbf{\color{orange}[description]}\\
Here is a prediction, between 0 and 100\%, for if \textbf{\color{orange}[condition]} happens by \textbf{\color{orange}[expiry]}
     }%
}

{\Large\textbf{\\ \\ \\Base Rate:\\}}

\fbox{%
  \parbox{5.3in}{%
You are trying to forecast if an event will take place in the future. For example, if the event is 'A member of the house of representatives steps down before the end of their term' the question might be 'How many times in the past 100 years has a member of the house of representatives stepped down before the end of their term.' If the event is 'Joe Biden dies in the next year' the question might be 'How likely is it than an 80 year old dies in the next year?' If the event is 'Company X has a profit of Y in 2022' the question might be 'What was the companies profits in each of the last 5 years?' Pose a question about the frequency of similar events that will act as a base rate for this event: \textbf{\color{orange}[name]} -- \textbf{\color{orange}[condition]}. "
     }%
}      
   
Return \textbf{\color{blue}[base rate question]}

\fbox{%
  \parbox{5.3in}{%
You are trying to forecast if an event will take place in the future: \textbf{\color{orange}[condition]}. Answer the following question: \textbf{\color{blue}[base rate question]} Give your response as a complete sentence.
     }%
}      

Return \textbf{\color{blue}[base rate]}

\fbox{%
  \parbox{5.3in}{%
[Forecaster Text]\\

Predict the likelihood of the following event by the given date.\\
You will make a prediction on the event called \textbf{\color{orange}[name]}.\\
The event will be true if the following condition is satisfied: \textbf{\color{orange}[condition]}\\
This condition must be satisfied by \textbf{\color{orange}[expiry]}\\
Currently the date is \textbf{\color{orange}[today]}, so the event must happen within the next \textbf{\color{orange}[number of days]} days.\\
Here is additional information about this event: \textbf{\color{orange}[description]}\\
Here a base rate for this event: \textbf{\color{blue}[base rate]} \\
Here is a prediction, between 0 and 100\%, for if \textbf{\color{orange}[condition]} happens by \textbf{\color{orange}[expiry]}
     }%
}

\textbf{\Large \\ \\ \\ Both Sides:\\}

\textit{Positive Prompt}:

\fbox{%
  \parbox{5.3in}{%
You are part of a team trying to figure out in an event is likely to take place.You are tasked with detailing all evidence that the event will happen. Describe everything we must consider that may make the event more likely to take place. This includes specific facts about the event that your team should consider, as well as sequences of events that may make the event more likely. For example, consider the event: Kamala Harris become president in 2024. You may want to mention her current role as vice president, as well as the possibility that Joe Biden may die or be too ill to run. Additionally, make use of base rates of similar situations. Using the above example this may include information such as the average 80 year old has about a 7\% chance of dying in the next year. Write a paragraph which you will share your your team detailing the evidence that may increase the likelihood of the event taking place. \\\\

You will be generating the supporting evidence for the event called \textbf{\color{orange}[name]}\\
The event will be true if the following condition is satisfied: \textbf{\color{orange}[condition]}\\
This condition must be satisfied by \textbf{\color{orange}[expiry]} \\
Currently the date is \textbf{\color{orange}[today]}, so the event must happen within the next \textbf{\color{orange}[number of days]} days. \\
Here is additional information about this event: \textbf{\color{orange}[description]} \\
Please give all the evidence that the condition will happen by \textbf{\color{orange}[expiry]}.
         }%
}

Return \textbf{\color{blue}[pros]}

\textit{Negative Prompt}:

\fbox{%
  \parbox{5.3in}{%
You are part of a team trying to figure out in an event is likely to take place. You are tasked with detailing all evidence that the event will not happen. Describe everything we must consider that may make the event less likely to take place. This includes specific facts about the event that your team should consider, as well as sequences of events that may make the event more less. For example, consider the event: Kamala Harris become president in 2024. You may want to mention that Joe Biden has only served one term. You may also want to mention her lack of popularity, so that even if Joe Biden does not run, she may not be chosen as the nominee. Additionally, make use of base rates of similar situations. Using the same example, this may include information such as only 5 presidents have chosen not to run for a second term.  Write a paragraph which you will share your your team detailing the evidence that may decrease the likelihood of the event taking place. \\\\
      
You will generating the evidence that suggests the following event does not take place: \textbf{\color{orange}[name]} \\
The event will be true if the following condition is satisfied: \textbf{\color{orange}[condition]}. \\
This condition must be satisfied by \textbf{\color{orange}[expiry]} \\
Currently the date is \textbf{\color{orange}[today]}, so the event must happen within the next \textbf{\color{orange}[number of days]} days. \\
Here is additional information about this event: \textbf{\color{orange}[description]} \\
Please give all the evidence that the condition will not happen by \textbf{\color{orange}[expiry]} or at all. \\
            }%
}    

Return \textbf{\color{blue}[cons]}

\fbox{%
  \parbox{5.3in}{%
[Forecaster Text]

Predict the likelihood of the following event by the given date.\\
You will make a prediction on the event called \textbf{\color{orange}[name]}.\\
The event will be true if the following condition is satisfied: \textbf{\color{orange}[condition]}\\
This condition must be satisfied by \textbf{\color{orange}[expiry]}\\
Currently the date is \textbf{\color{orange}[today]}, so the event must happen within the next \textbf{\color{orange}[number of days]} days.\\
Here is additional information about this event: \textbf{\color{orange}[description]}\\
Here is argument for why the event may come true:  \textbf{\color{blue}[pros]} \\
Here is argument for why the event may not come true:  \textbf{\color{blue}[cons]} \\
Here is a prediction, between 0 and 100\%, for if \textbf{\color{orange}[condition]} happens by \textbf{\color{orange}[expiry]}
     }%
}

\textbf{\Large \\ \\ \\ Sequences:\\}

\fbox{%
  \parbox{5.3in}{%
You are a forecaster tasked with generating all the different possible sequences of events that would cause an event to happen. \\\\

[EVENT] Kamala Harris is the Democratic Nominee for 2024 \\\relax
[CURRENT DATE] 2022-8-1 \\\relax
[EXPIRY DATE] 2024-11-5 \\\relax
[ADDITIONAL INFORMATION] Joe Biden has only served on term, but is 80 years old. \\\relax
[QUESTIONS] Will the event 'Kamala Harris is the Democratic Nominee for 2024' happen by 2024-11-5? \\\relax
[POTENTIAL INCITING EVENTS] Joe Biden dies \\\relax
[PATH TO POSITIVE OUTCOME] \\\relax
1. Joe Biden decides not to run for a second term \\\relax
2. Kamala Harris is selected as the Democratic Nominee \\\relax
OUTCOME ACHIEVED: Kamala Harris is the Democratic Nominee for 2024 \\\relax
[PATH TO POSITIVE OUTCOME] \\\relax
1. Joe Biden dies in office \\\relax
2. Kamala Harris becomes incumbent president \\\relax
3. Kamala Harris is selected as Democratic Nominee \\\relax
OUTCOME ACHIEVED: Kamala Harris is the Democratic Nominee for 2024 \\\relax
END \\\\\relax
    
[EVENT] A human sets foot on Mars \\\relax
[CURRENT DATE] 2022-8-1 \\\relax
[EXPIRY DATE] 2030-01-01 \\\relax
[ADDITIONAL INFORMATION] On Dec 8, 2020, Musk claimed a human setting foot on Mars was achievable in 6 years, and 4 years if we get lucky. The Starship rocket that Musk envisions would deliver this vision has made excellent progress recently. \\\relax
[QUESTIONS] Will the event 'A human sets foot on Mars' happen by 2030-01-01? \\\relax
[POTENTIAL INCITING EVENTS] None \\\relax
[PATH TO POSITIVE OUTCOME] \\\relax
1. SpaceX lands a human on Mars\\ \relax
OUTCOME ACHIEVED:  A human sets foot on Mars \\\relax
[PATH TO POSITIVE OUTCOME] \\\relax
1. NASA lands a human on Mars \\\relax
OUTCOME ACHIEVED: A human sets foot on Mars \\\relax
[PATH TO POSITIVE OUTCOME] \\\relax
1. A non-US country lands a human on Mars \\\relax
OUTCOME ACHIEVED: A human sets foot on Mars \\\relax
END \\\\\relax
    
[EVENT] \textbf{\color{orange}[condition]} \\\relax
[CURRENT DATE] \textbf{\color{orange}[today]} \\\relax
[EXPIRY DATE] \textbf{\color{orange}[expiry]} \\\relax
[ADDITIONAL INFORMATION] \textbf{\color{orange}[description]} \\\relax
[QUESTIONS] Will the event '\textbf{\color{orange}[condition]}' happen by \textbf{\color{orange}[expiry]}? \\\relax
[POTENTIAL INCITING EVENTS]

     }%
} 

Return \textbf{\color{blue}[Positive Sequences]}
 
   \fbox{%
  \parbox{5.3in}{%
You are a forecaster tasked with generating all the different possible sequences of events that would cause an event to NOT happen.\\\\\relax

[EVENT] Kamala Harris is the Democratic Nominee for 2024 \\\relax
[OPPOSITE] Kamala Harris is not the Democratic Nominee for 2024 \\\relax
[END]\\\\\relax
  
[EVENT] A human sets foot on Mars by 2030-01-01 \\\relax
[OPPOSITE] A human does not set foot on Mars by 2030-01-01 \\\relax
[END] \\ \\\relax
  
[EVENT] \textbf{\color{orange}[condition]}

     }%
} 

Return \textbf{\color{blue}[Opposite Event]}

\fbox{%
  \parbox{5.3in}{%
You are a forecaster tasked with generating all the different possible sequences of events that would cause an event to NOT happen. \\\\\relax

[EVENT] Kamala Harris is the Democratic Nominee for 2024 \\\relax
[EVENT OPPOSITE] Kamala Harris is not the Democratic Nominee for 2024 \\\relax
[CURRENT DATE] 2022-8-1 \\\relax
[EXPIRY DATE] 2024-11-5 \\\relax
[ADDITIONAL INFORMATION]. \\\relax
[QUESTIONS] Will the event 'Kamala Harris is the Democratic Nominee for 2024'
NOT happen by 2024-11-5? \\\relax
[POTENTIAL INHIBITING EVENTS] Joe biden decides to run for a second term. \\\relax
[PATH TO NEGATIVE OUTCOME] \\\relax
1. Joe Biden decides to run for a second term \\\relax
2. Joe Biden is selected as the Democratic Nominee \\\relax
OUTCOME NOT ACHIEVED: Kamala Harris is NOT the Democratic Nominee for 2024 \\\relax
[PATH TO NEGATIVE OUTCOME] \\\relax
1. Joe Biden dies in office \\\relax
2. Kamala Harris becomes incumbent president \\\relax
3. Another democrat, such as Gavin Newsom, is selected as the democratic nominee
OUTCOME NOT ACHIEVED: Kamala Harris is NOT the Democratic Nominee for 2024 \\\relax
[PATH TO NEGATIVE OUTCOME] \\\relax
1. Joe Biden dies decides not to run for a second term \\\relax
2. Another democrat, such as Gavin Newsom, is selected as the democratic nominee \\\relax
OUTCOME NOT ACHIEVED: Kamala Harris is NOT the Democratic Nominee for 2024 \\\relax
[PATH TO NEGATIVE OUTCOME] \\\relax
1. Kamala Harris does not run for President in 2024 \\\relax
OUTCOME NOT ACHIEVED: Kamala Harris is NOT the Democratic Nominee for 2024 \\\relax
END \\\relax
    
[EVENT] A human sets foot on Mars by 2030-01-01 \\\relax
[EVENT OPPOSITE] A human does not foot on Mars by 2030-01-01 \\\relax
[CURRENT DATE] 2022-8-1 \\\relax
[EXPIRY DATE] 2030-01-01 \\\relax
[ADDITIONAL INFORMATION] On Dec 8, 2020, Musk claimed a human setting foot on Mars was achievable in 6 years, and 4 years if we get lucky. The Starship rocket that Musk envisions would deliver this vision has made excellent progress recently. \\\relax
[QUESTIONS] Will the event 'A human sets foot on Mars'"\\\relax
NOT happen by 2030-01-01? 
[POTENTIAL INHIBITING EVENTS] None \\\relax
[PATH TO NEGATIVE OUTCOME]     \\\relax
1. SpaceX lands a human on Mars after 2030-01-01 \\\relax
OUTCOME NOT ACHIEVED:  A human does NOT set foot on Mars by 2030-01-01 \\\relax
[PATH TO NEGATIVE OUTCOME] \\\relax
1. SpaceX decides to cancel the project to land a human on Mars \\\relax
OUTCOME NOT ACHIEVED:  A human does NOT set foot on Mars by 2030-01-01 \\\relax
END \\\relax
    
[EVENT] \textbf{\color{orange}[condition]} by \textbf{\color{orange}[expiry]} \\\relax
[EVENT OPPOSITE] \textbf{\color{orange}[Opposite Event]} by \textbf{\color{orange}[expiry]} \\\relax
[CURRENT DATE] \textbf{\color{orange}[today]} \\\relax
[EXPIRY DATE] \textbf{\color{orange}[expiry]} \\\relax
[ADDITIONAL INFORMATION] \textbf{\color{orange}[description]} \\\relax
[QUESTIONS] Will the event '\textbf{\color{orange}[condition]}' \\\relax
 NOT happen by \textbf{\color{orange}[expiry]}?
[POTENTIAL INHIBITING EVENTS] 
     }%
} 

Return \textbf{\color{blue}[Negative Sequences]}

\fbox{%
  \parbox{5.3in}{%
[Forecaster Text]\\

You will make a prediction on the event called \textbf{\color{orange}[name]} \\\relax
The event will be true if the following condition is satisfied: \textbf{\color{orange}[condition]} \\\relax
This condition must be satisfied by \textbf{\color{orange}[expiry]} \\\relax
Currently the date is \textbf{\color{orange}[today]} \\\relax
Here is additional information about this event: \textbf{\color{orange}[description]} \\\relax

Here are a number sequences that may lead to this event happening: \\
Potential Sequence \textbf{\color{blue}[i]} : \\
\textbf{\color{blue}[Positive Sequence i]} \\\\

Here are a number sequences that may lead to this event not happening:\\
Potential Sequence \textbf{\color{blue}[i]}: \\
\textbf{\color{blue}[Negative Sequence i]} \\\\
    
Now talk through your rationale, including all possible sequences and that may lead to this event. Please take into account that to resolve as true, the event must happen in the next \textbf{\color{orange}[number of days]} days. After considering all the information, output a final probability between 0 and 1 for if the event \textbf{\color{orange}[condition]}, will happen between \textbf{\color{orange}[today]} and \textbf{\color{orange}[expiry]}.
     }%
}

\textbf{\Large \\ \\ \\ Crowd:\\}

\fbox{%
  \parbox{5.3in}{%
  You must ask an expert to make a prediction on a single event. Your job is to decide which expert you would like to  ask. For example, if you were to pick an expert to predict if SpaceX will land on Mars, you may choose to ask an aerospace engineer. You must pick an expert to make a prediction on if the following event will take place: \textbf{\color{orange}[condition]}. Here is additional information on the event: \textbf{\color{orange}[description]}. Who would you like to ask? I choose to talk to 
         }%
} 

return \textbf{\color{blue}[job]}

\fbox{%
  \parbox{5.3in}{%
You are \textbf{\color{blue}[job]}. Using your expertise, you will make a prediction on the likelihood of an event taking place. Using your background knowledge, consider past events, and use this to inform your prediction about the future.\\

You will make a prediction on the event called \textbf{\color{orange}[name]} \\
The event will be true if the following condition is satisfied: \textbf{\color{orange}[condition]}  \\
This condition must be satisfied by \textbf{\color{orange}[expiry]} \\
Currently the date is \textbf{\color{orange}[today]} \\
Here is additional information about this event: \textbf{\color{orange}[description]}  \\
      
Now talk through your rationale. Begin by discussing you own expertise. Next talk through the specific details about the event. You will give two probabilities. First, output the probability, between 0 and 1, that \textbf{\color{orange}[condition]} will happen ever. Then, output the probability, between 0 and 1, that \textbf{\color{orange}[condition]} will happen between \textbf{\color{orange}[today]} and \textbf{\color{orange}[expiry]}, so in the next \textbf{\color{orange}[number of days]} days.'
         }%
}

\textbf{\Large \\ \\ \\ External News:\\}

\fbox{%
  \parbox{5.3in}{%
You are summarizing an event into just the main words involved, in order to best search for the event in a search engine. You may name people, places, events or anything else. For example, the main entities involved in SpaceX lands human on Mars, would  be: \\
*SpaceX\\
*Mars\\ \\
the main entities involved in Kamala Harris Wins Democratic Nomination are: \\
*Kamala Harris\\
*Democratic Nomination\\

 What are the primary entities involved in this event: \textbf{\color{orange}[name]} (\textbf{\color{orange}[condition]})?
 Give \textbf{\color{orange}[number of terms]} terms: \\*
           }%
}

Return \textbf{\color{blue}[search terms]}\\
\textbf{\color{blue}[Hackernews headlines]} = query-HackerNews-API(\textbf{\color{blue}[search terms]}) \\
\textbf{\color{blue}[NYT headlines]} = query-NYT-API(\textbf{\color{blue}[search terms]}) \\

  \fbox{%
  \parbox{5.3in}{%
You are examining some newspaper headlines to try to predict if an event will take place. Read through the headlines and remove any which are totally irrelevant to the given topic. Additionally, remove headlines if they are identical to an existing headline. \\\\\relax
            
\textbf{\color{blue}[Hackernews headlines]} \\ \\
            
You are interested any headlines that could lead to the event or indicate that it will not happen: \textbf{\color{orange}[name]} Keep any headlines that might indicate that the event will take place. Additionally, keep any headlines that might indicate that the event will not take place. Remove irrelevant and duplicate headlines. If none of the headlines are very relevant, output NONE. 
                   }%
}    
     
Return \textbf{\color{blue}[filtered Hackernews headlines]}

  \fbox{%
  \parbox{5.3in}{%

You are examining some newspaper headlines to try to gather information about an event. Many of the headlines contain irrelevant information, but some headlines may contain valuable pieces of information. Read through the headlines and look for information which directly relates to the event you are studying. Then return a list of all relevant information extracted from the headlines. You may return entire headlines, small pieces of headlines, or paraphrased headlines. For example, you may be looking through headlines that that contain evidence about if Kamala Harris will be the next president. If a headline says 'Headline 1 -- 2023-01-01: Kamala Harris states in press conference on Tuesday that she accepts nomination and looks forward to leading the Country' you may return '2023-01-01: Kamala Harris says she accepts nomination.You are looking for information about if the following event will take place: \textbf{\color{orange}[name]} (\textbf{\color{orange}[condition]}). Here are the headlines you have access to: [HEADLINES]:" \\\\\relax
            
\textbf{\color{blue}[NYT headlines]} \\

Please pull all information from the headlines that may directly increase or decrease the likelihood of the event: \textbf{\color{orange}[name]} (\textbf{\color{orange}[condition]}). Ignore any headlines that do not directly contain information about the exact event: \textbf{\color{orange}[name]} (\textbf{\color{orange}[condition]}). If none of the headline are very relevant, output NONE. \\\\
                               }%
} 
return \textbf{\color{blue}[filtered NYT headlines]}\\
\fbox{%
  \parbox{5.3in}{%
Read through each of the headlines and paraphrase each one as it relates to predicting the likelihood of the following event: \textbf{\color{orange}[name]} (\textbf{\color{orange}[condition]}). Include the date of the headline at the beginning. Do not include any interpretation of the headlines, only paraphrasing.

\textbf{\color{blue}[filtered NYT headlines]}
            
                               }%
}    

Return \textbf{\color{blue}[summarized NYT headlines]}\\

  \fbox{%
  \parbox{5.3in}{%
            [Forecaster Text]\\
            
You will make a prediction on the event called  \textbf{\color{orange}[name]} \\
The event will be true if the following condition is satisfied:  \textbf{\color{orange}[condition]} \\
This condition must be satisfied by  \textbf{\color{orange}[expiry]} \\
Currently the date is  \textbf{\color{orange}[today]}, so the event must happen within the next  \textbf{\color{orange}[number of days]} days. \\
Here is additional information about this event:  \textbf{\color{orange}[description]} \\\\

Here are recent headlines from Hackernews on the topic. \\\relax
\textbf{\color{blue}[filtered Hackernews headlines]} \\
  
 Here are recent headlines from the New York Times on the topic. \\\relax
\textbf{\color{blue}[summarized NYT headlines]} \\

 Based on all the information available to you how likely do you think it is that this event has takes place by  \textbf{\color{orange}[expiry]}? Give your explanation and probability.
      
                                     }%
}

\newpage
\section{Example Events from GleanGen}
\label{moreexamples}

In this Appendix, we provide examples of events from each of the four analyzed categories (Covid-19, Finance, Technology Industry, and Misc.) in the GleanGen dataset. These include two negatively resolving examples (Resolution: Condition Not Met) and two positive examples (Resolution: Condition Met).  

\begin{figure}[h]
\includegraphics[scale=0.8]{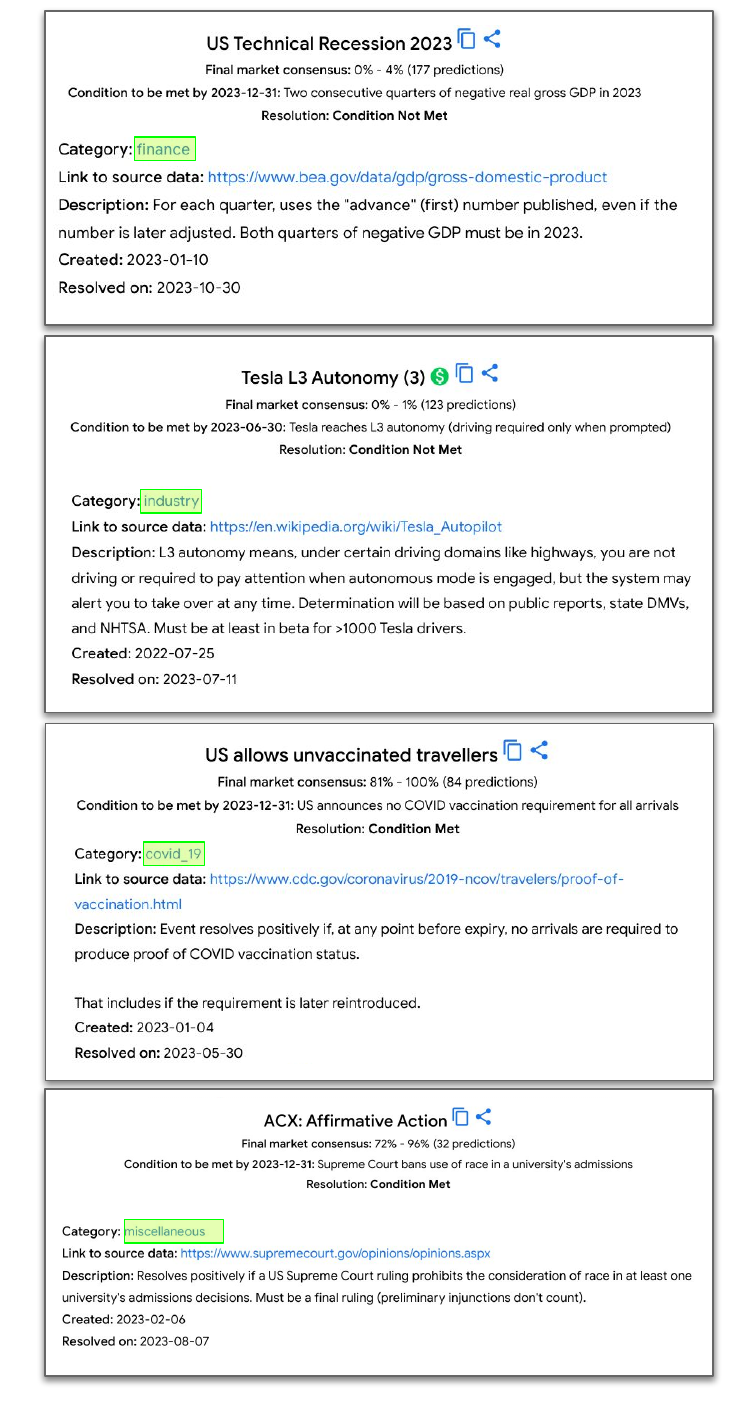}
\end{figure}

\newpage
\section{Larger Figure \ref{just_answer}}
\label{larger1}

\begin{figure}[h]
\includegraphics[scale=0.7]{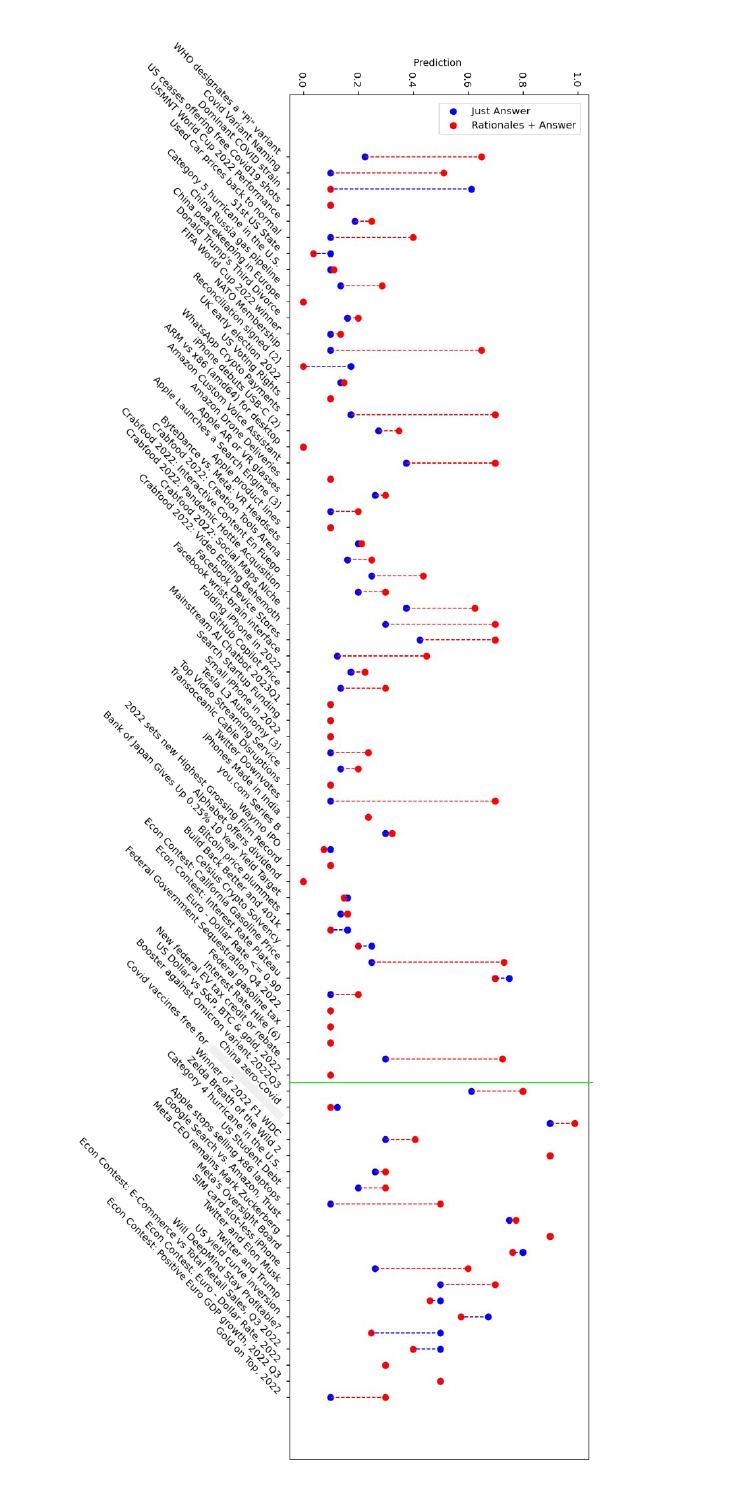}
\caption{Larger scale version of Figure \ref{just_answer}. Figure demonstrates that predicted probabilities are consistently higher when the model is prompted to first produce a rationale. Events which ultimately resolve as True are below the green line and events which resolve as False are above the green line. }
\label{larger}
\end{figure}

\newpage

\end{document}